\documentclass[a4paper,conference]{IEEEtran}

\usepackage{graphicx}
\usepackage{CJKutf8}
\usepackage{comment}
\usepackage{multirow}
\usepackage{url}
\usepackage{framed}

\begin{document}

\title{Survey on Deep Learning-based\\Kuzushiji Recognition}

\author{\IEEEauthorblockN{Kazuya Ueki}
\IEEEauthorblockA{School of Information Science\\Meisei University\\
Email: kazuya.ueki@meisei-u.ac.jp}
\and
\IEEEauthorblockN{Tomoka Kojima}
\IEEEauthorblockA{School of Information Science\\
Meisei University\\
Email: 18j5061@stu.meisei-u.ac.jp}}

\maketitle

\begin{abstract}
Owing to the overwhelming accuracy of the deep learning method demonstrated at the 2012 image classification competition, deep learning has been successfully applied to a variety of other tasks.
The high-precision detection and recognition of Kuzushiji, a Japanese cursive script used for transcribing historical documents, has been made possible through the use of deep learning.
In recent years, competitions on Kuzushiji recognition have been held, and many researchers have proposed various recognition methods.
This study examines recent research trends, current problems, and future prospects in Kuzushiji recognition using deep learning.
\end{abstract}


%
\IEEEpeerreviewmaketitle

\section{Introduction}

Kuzushiji has been commonly used in Japan for more than a thousand years.
However, since the 1900s, schools have no longer been teaching Kuzushiji, and only a few thousand people in Japan can currently read and understand it.
Hiragana characters\footnote{Hiragana is one of the three different character sets used in Japanese writing. Each Hiragana character represents a particular syllable. There are 46 basic characters.} have a root Kanji\footnote{Kanji is another one of the three character sets used in the Japanese language. Along with syllabaries, Kanji is made up of ideographic characters, and each letter symbolizes a specific meaning. 
Most Kanji characters were imported from China, although some were developed in Japan. 
Although there are approximately 50,000 Kanji characters, only approximately 2,500 are actually used in daily life in Japan.} called a Jibo\footnote{A Jibo is a root Kanji character of Hiragana. For example, the character ``\begin{CJK}{UTF8}{ipxm}あ\end{CJK}'' is derived from different Jibos including ``\begin{CJK}{UTF8}{ipxm}安\end{CJK}'' and ``\begin{CJK}{UTF8}{ipxm}阿\end{CJK}.''}, leading to various shapes for a single character; training is required to read characters that differ from modern Hiragana.
For this reason, many researchers have been working on Kuzushiji recognition using machine learning techniques.
Recently, with the advent of deep learning, research on Kuzushiji recognition has accelerated and the accuracy of the methods has significantly improved.
In this paper, we present a survey and analysis of recent methods of Kuzushiji recognition based on deep learning.

\section{Representative Research on Kuzushiji Recognition}
\label{sec:previous_work}

Many studies on Kuzushiji recognition were conducted prior to the introduction of deep learning.
The ``Historical Character Recognition  Project'' \cite{Yamada2001}, which was initiated in 1999, reported the development of a system to support the transcription of ancient documents.
In this project, to develop a historical document research support system, the authors studied a character database, corpus, character segmentation, character recognition, intellectual transcription support system, and a digital dictionary.
Specifically, they developed a computerized historical character dictionary using stroke information \cite{Yamada2002} as well as Japanese off-line hand-written optical character recognition (OCR)  technology, and implemented a Kuzushiji recognition system for 67,739 categories by combining on-line and off-line recognition methods \cite{Onuma2007}. 
Other methods, such as \cite{Horiuchi2011}\cite{Kato2014}, 
which is a recognition method using self-organizing maps, and \cite{Hayasaka2015}, which is a recognition method using a neocognitron, have also been proposed.

Since the introduction of deep learning, further research on Kuzushiji recognition has become increasingly active, and various methods have been proposed. 
During the early introductory stage of deep learning, most recognition methods \cite{Hayasaka2016}\cite{Ueda2018} were based on approximately 50 different Hiragana images with recognition rates of 75\% to 90\%.
There were also studies in which more than 1,000 characters including Hiragana, and Katakana\footnote{In the same way as Hiragana, Katakana is one of the three different character sets used in Japanese. Katakana is also a phonetic syllabary, in which each letter represents the sound of a syllable. There are also 46 basic characters.} and Kanji, were recognized \cite{Tomoka2019}, along with the results of character recognition in documents of the Taiwan Governor's Office, which dealt with more than 3,000 characters \cite{Yang2019}.
In these studies, the problems of large numbers of classes, an unbalanced number of data between classes, and a variation of characters were solved through a data augmentation commonly used in deep learning training.

In addition, a network that outputs a three-character string has also been proposed as a method for recognizing consecutive characters \cite{Nagai2017}.
This method uses single-character and binary classifiers to distinguish between characters; the character strings are then recognized using bidirectional long short-term memory (BLSTM).
The authors reported that the recognition rate of a single character was approximately 92\%; however, the recognition rate of three characters was only approximately 76\%.
Similarly, a method for recognizing a string of three consecutive characters using a sliding window and BLSTM was proposed \cite{Ueki2020}.
The authors used the tendency in which the maximum output probability of a neural network is not particularly high for a misaligned character image but is high for an accurately aligned image, and increased the recognition rate to 86\% by integrating multiple results. 
In addition, a deep learning method for recognizing a series of Kuzushiji phrases using an image from ``The Tale of Genji'' was proposed \cite{Genji2019}.
An end-to-end method with an attention mechanism was applied to recognize consecutive Kuzushiji characters within phrases.
This method can recognize phrases written in Hiragana (47 different characters) with 78.92\% accuracy, and phrases containing both Kanji (63 different characters) and Hiragana with 59.80\% accuracy.

In recent years, research on Kuzushiji recognition has become increasingly active since the Kuzushiji dataset first became publicly available \cite{Kitamoto2019inpact}. 
With the development of this database, a PRMU algorithm contest described in \ref{ssec:PRMU} and a Kaggle competition introduced in \ref{ssec:Kaggle}  were held, and many researchers have started to work on Kuzushiji recognition.
The preparation, progress, and results of a Kuzushiji recognition competition, knowledge obtained from the competition, and the value of utilizing machine learning competitions have also been reported \cite{Kitamoto2019}\cite{Kitamoto2020}.
In these reports, the results of the Kuzushiji recognition competition showed that existing object detection algorithms such as a Faster R-CNN \cite{Faster R-CNN} and cascade R-CNN \cite{Cascade R-CNN} are also effective for Kuzushiji detection.
At the forefront of Kuzushiji recognition, end-to-end approaches for actual transcriptions are becoming the mainstream.
As a representative method, an end-to-end method, KuroNet, was proposed to recognize whole pages of text using the U-Net architecture \cite{KuroNet}\cite{KuroNet2020}.
The authors demonstrated that KuroNet can handle long-range contexts, large vocabularies, and non-standardized character layouts by predicting the location and identity of all characters given a page of text without any preprocessing.
To recognize multiple lines of historical documents, a document reading system inspired by human eye movements was proposed, and the results of evaluations of the PRMU algorithm contest database described in \ref{ssec:PRMU} \cite{Anh2017} and the Kaggle competition database described in \ref{ssec:Kaggle} \cite{Anh2020} were reported.
In addition, a two-dimensional context box proposal network used to detect Kuzushiji in historical documents was proposed \cite{Anh2019}.
The authors employed VGG16 to extract features from an input image and BLSTM \cite{BLSTM} for exploring the vertical and horizontal dimensions, and then predicted the bounding boxes from the output of the two-dimensional context.

In a study on the practical aspects of transcription, a new type of OCR technology was proposed to reduce the labor of a high-load transcription \cite{Yamamoto2016}. 
This technology is not a fully automated process but aims to save labor by dividing tasks between experts and non-experts and applying an automatic processing.
The authors stated that a translation can be made quickly and accurately by not aiming at a decoding accuracy of 100\% using only the OCR of automatic processing, but by leaving the characters with a low degree of certainty as a ``\begin{CJK}{UTF8}{ipxm}〓\end{CJK} ({\it geta})'' and entrusting the evaluation to experts in the following process.
A method for automatically determining which characters should be held for evaluation using a machine learning technique was also proposed \cite{Tomoka2020}\cite{ICPRAM2020}.  
This method can automatically identify difficult-to-recognize characters or characters that were not used during training based on the confidence level obtained from the neural network. 

A study on an interface of a Kuzushiji recognition system also introduced a system that can operate on a Raspberry Pi 
without problems and with almost the same processing time and accuracy as in previous studies; there was also no need for a high-performance computer \cite{Takeuti2019}.

As another area of focus, a framework for assisting humans in reading Japanese historical manuscripts, formulated as a constraint satisfaction problem, and a system for transcribing Kuzushiji and its graphical user interface have also been introduced \cite{Sando2018}\cite{Yamazaki2018}.

An interactive system was also been proposed to assist in the transcription of digitized Japanese historical woodblock-printed books \cite{Ritsumeikan2017}. 
This system includes a layout analysis, character segmentation, transcription, and the generation of a character image database.
The procedures for applying the system consist of two major phases.
During the first phase, the system automatically produces provisional character segmentation data, and users
interactively edit and transcribe the data into text data for storage in the character image database.
During the second phase, the system conducts automatic character segmentation and transcription using the database generated during the first phase.
By repeating the first and second phases with a variety of materials, the contents of the character image database can be enhanced and the performance of the system in terms of character segmentation and transcription will increase.

\section{Datasets}

\subsection{Kuzushiji Dataset}
\label{ssec:CODH}

\begin{figure}[t]
\centering
\includegraphics[height=60mm]{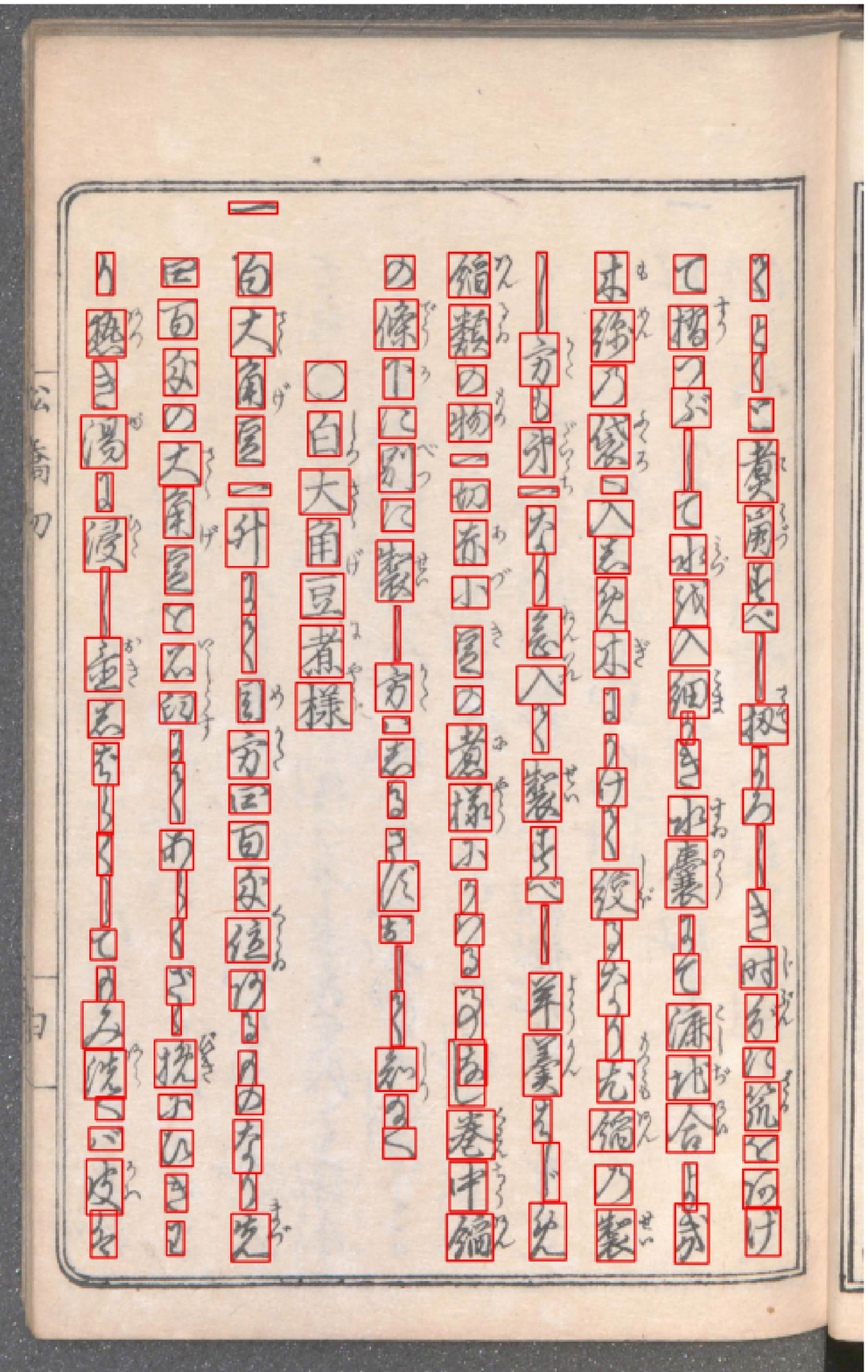}
\includegraphics[height=60mm]{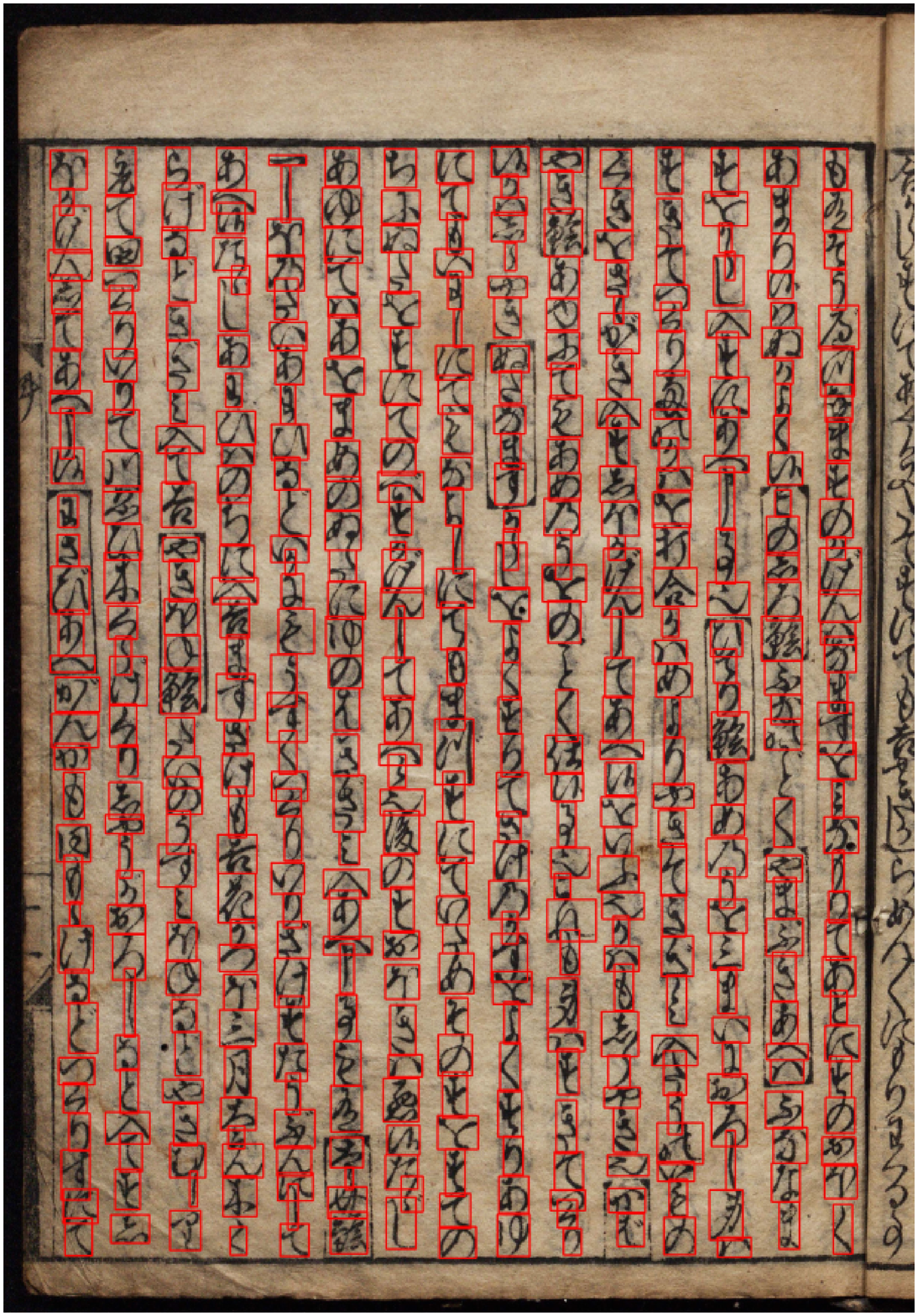}
\caption{Example images of Kuzushiji dataset}
\label{fig:KuzushijiDB}
\end{figure}

The Kuzushiji dataset\footnote{\url{http://codh.rois.ac.jp/char-shape/}} consists of 6,151 pages of image data of 44 classical books held by National Institute of Japanese Literature and published by ROIS-DS Center for Open Data in the Humanities (CODH).
Example images included in the Kuzushiji dataset are shown in Fig. \ref{fig:KuzushijiDB}.
This Kuzushiji database comprises bounding boxes for characters, including 4,328 character types and 1,086,326 characters.
There is a large bias in the number of data depending on the class: 
The class with the largest number of images is ``\begin{CJK}{UTF8}{ipxm}の\end{CJK}'' (character code, U+306B), which has 41,293 images; many classes have extremely few images, and 790 classes only have 1 image.

Kuzushiji-MNIST, Kuzushiji-49, and Kuzushiji-Kanji were also provided as a subset of the above dataset\footnote{\url{https://github.com/rois-codh/kmnist}} \cite{Clanuwat2018}.
These datasets not only serve as a benchmark for advanced classification algorithms, they can also be used in more creative areas such as generative modeling, adversarial examples, few-shot learning, transfer learning, and domain adaptation.
Kuzushiji-MNIST has 70,000 28$\times$28 grayscale images with 10 Hiragana character classes.
Kuzushiji-49 is an imbalanced dataset that has 49 classes (28$\times$28 grayscale, 270,912 images) containing 48 Hiragana characters and one Hiragana iteration mark.
Kuzushiji-Kanji is an imbalanced dataset with a total of 3,832 Kanji characters (64$\times$64 grayscale, 140,426 images), ranging from 1,766 examples to only a single example per class.

\subsection{Electronic Kuzushiji Dictionary Database}
\label{ssec:TokyoUniv}

The Electronic Kuzushiji Dictionary Database\footnote{\url{https://wwwap.hi.u-tokyo.ac.jp/ships/shipscontroller}} is a database of glyphs and fonts collected from ancient documents and records in the Historiographical Institute of the University of Tokyo; it contains approximately 6,000 different characters, 2,600 different vocabularies, and 280,000 character image files.
This database contains character forms from various periods from the Nara period (8th century) to the Edo period (18th century).

\subsection{Wooden Tablet Database}

\begin{figure}[t]
\centering
\begin{framed}
\includegraphics[height=55mm]{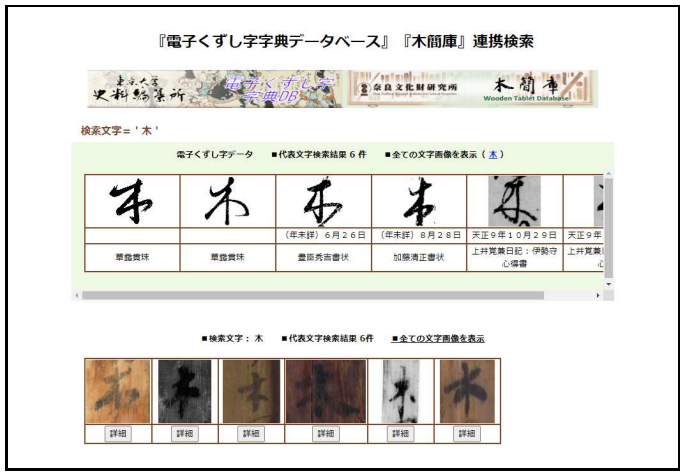}
\end{framed}
\caption{``Electronic Kuzushiji dictionary database'' and ``wooden tablet database'' collaborative search function}
\label{fig:collaborative_search}
\end{figure}

The Nara National Research Institute for Cultural Properties has developed and published a database that collects images of glyphs and fonts allowing the recognition of inscriptions written on wooden blocks excavated from underground sites.
The database contains approximately 24,000 characters, 1,500 character types, and 35,000 character images.
It contains information from the Asuka-Nara period, which is not often included in the Electronic Kuzushiji Dictionary Database described in \ref{ssec:TokyoUniv}.
For this reason, the ``Electronic Kuzushiji Dictionary Database'' and ``wooden tablet database'' collaborative search function\footnote{\url{http://clioapi.hi.u-tokyo.ac.jp/ships/ZClient/W34/z_srch.php}} shown in Fig. \ref{fig:collaborative_search}, which integrates the two databases, was provided for convenience.

\section{Benchmarks}

\subsection{PRMU Algorithm Contest}
\label{ssec:PRMU}

The tasks used in the 2017 and 2019 PRMU Algorithm Contest\footnote{A contest held annually by the Pattern Recognition and Media Understanding (PRMU) for the purpose of revitalizing research group activities.} required recognizing Kuzushiji contained in the designated region of an image of a classical Japanese book and outputting the Unicode of each character.
In this contest, a total of 46 types of Hiragana characters that do not include Katakana or Kanji needed to be recognized.
The 2017 contest had three tasks for three different difficulty levels, levels 1, 2, and 3, depending on the number of characters contained in the rectangle. 
The participants were required to recognize single segmented characters in level 1, three consecutive characters in the vertical direction in level 2, and three or more characters in the vertical and horizontal directions in level 3.
In the 2019 contest, the task was to recognize three consecutive characters in the vertical direction as in level 2 in 2017.

Herein, we introduce the tasks of the 2017 contest and the methods used by the best performing teams \cite{alcon2017}.
The dataset applied in 2017 was constructed from 2,222 scanned pages of 15 historical books provided by CODH. 
One of the 15 books was used as the test data because it contained many fragmented and noisy patterns, as well as various backgrounds.
The dataset for level 1 consisted of 228,334 single Hiragana images, and test data of 47,950 were selected from the dataset. 
To improve the accuracy, multiple models were trained and a voting-based ensemble method was employed to integrate the results of many different models. 
The level 2 dataset consists of 92,813 images of three consecutive characters, and a test set of 13,648 images was selected from the dataset. 
A combined architecture of a CNN , BLSTM, and connectionist temporal classification (CTC) was employed \cite{CTC2016}, and a sequence error rate (SER) of 31.60\% was achieved.
In the level 3 dataset, there are 12,583 images from which a test set of 1,340 images were selected.  
The authors employed a combination of vertical line segmentation and multiple line concatenation before applying a deep convolutional recurrent network.
The SER is 82.57\%, and there is still a significant need for improvement.

Another report evaluated a historical document recognition system inspired by human eye movements using the dataset from the 2017 PRMU algorithm contest \cite{Anh2017}.
This system includes two modules: a CNN for feature extraction and an LSTM decoder with an attention model for generating the target characters.
The authors achieved SERs of 9.87\% and 53.81\% at levels 2 and 3 of the dataset, respectively.

Now, we introduce the methods of the first through third place teams in the 2019 contest.
The dataset used in 2019 was also composed of 48 Hiragana character images cropped from the books in the Kuzushiji dataset provided by CODH.
Single-character images and images containing three consecutive characters in the vertical direction were provided as data for training.
A total of 388,146 single-character images were applied, and 119,997 images with three consecutive characters were used for training and 16,387 were used for testing. 

The first place team adopted a method dividing the characters into three images through a preprocessing, inputting each image into the CNN to extract the features, and recognizing three consecutive characters using two layers of a bidirectional gated recurrent unit (GRU) \cite{GRU}\footnote{\url{https://github.com/katsura-jp/alcon23}}. 
They achieved a rate of 90.63\% through a combination of three backbone models (SE-ResNeXt, DenseNet, and Inception-v4) . 
As data augmentation methods, in addition to a random crop and a random shift, the division position was randomly shifted up and down slightly during training for robustness to the division position.

The second place team used a CNN, BLSTM, and CTC in the first step and output three characters by majority voting during the second step.
In the first step, a CNN (six layers) was used to extract the features, BLSTM (two layers) was used to convert the features into sequential data, and CTC was used to output the text. 
To improve the accuracy, data augmentation such as a random rotation, random zoom, parallel shift, random noise, and random erasing \cite{Random Erasing} were used.

The third place team employed an algorithm that applies multi-label image classification.
In the first step, a multi-label estimation was conducted using an image classification model, and three characters were estimated in no particular order. 
During the second step, Grad-CAM \cite{Grad-CAM} identified and aligned the region of interest for each candidate character and output three consecutive characters.

\subsection{Kaggle Competition}
\label{ssec:Kaggle}

\begin{table*}[t]
\centering
\caption{Explanation and program implementation of the winning method in the Kaggle competition}
\small
\label{Table:Kaggle}
\begin{tabular}{c|c|l} \hline
 Rank & F value & URL \\
 \hline \hline 
 \multirow{2}{*}{1} & \multirow{2}{*}{0.950} & Explanation: https://www.kaggle.com/c/kuzushiji-recognition/discussion/112788  \\
 & & Implementation: https://github.com/tascj/kaggle-kuzushiji-recognition \\
 \hline
 \multirow{2}{*}{2} & \multirow{2}{*}{0.950} & Explanation: https://www.kaggle.com/c/kuzushiji-recognition/discussion/112712 \\
 & & Implementation: https://github.com/lopuhin/kaggle-kuzushiji-2019 \\
 \hline
 \multirow{2}{*}{3} & \multirow{2}{*}{0.944} & Explanation: https://www.kaggle.com/c/kuzushiji-recognition/discussion/113049 \\
 & & Implementation: https://github.com/knjcode/kaggle-kuzushiji-recognition-2019 \\
 \hline
 \multirow{2}{*}{4} & \multirow{2}{*}{0.942} & Explanation: https://www.kaggle.com/c/kuzushiji-recognition/discussion/114764  \\
 & & Implementation: https://github.com/linhuifj/kaggle-kuzushiji-recognition \\
 \hline
 \multirow{2}{*}{5} & \multirow{2}{*}{0.940} & Explanation: https://www.kaggle.com/c/kuzushiji-recognition/discussion/112771  \\
 & & Implementation: https://github.com/see--/kuzushiji-recognition \\
 \hline
 \multirow{2}{*}{7} & \multirow{2}{*}{0.934} & Explanation: https://www.kaggle.com/c/kuzushiji-recognition/discussion/112899  \\
 & & Implementation: https://www.kaggle.com/kmat2019/centernet-keypoint-detector \\
 \hline
 \multirow{2}{*}{8} & \multirow{2}{*}{0.920} & Explanation: https://www.kaggle.com/c/kuzushiji-recognition/discussion/113419  \\
 & & Implementation: https://github.com/t-hanya/kuzushiji-recognition \\
 \hline
 \multirow{2}{*}{9} & \multirow{2}{*}{0.910} & Explanation: https://www.kaggle.com/c/kuzushiji-recognition/discussion/112807  \\
 & & Implementation: https://github.com/mv-lab/kuzushiji-recognition \\
 \hline
 \multirow{2}{*}{13} & \multirow{2}{*}{0.901} & Explanation: https://www.kaggle.com/c/kuzushiji-recognition/discussion/113518  \\
 & & Implementation: https://github.com/jday96314/Kuzushiji \\
 \hline
 \multirow{2}{*}{15} & \multirow{2}{*}{0.900} & Explanation: https://www.kaggle.com/c/kuzushiji-recognition/discussion/114120  \\
 & & Implementation: https://github.com/statsu1990/kuzushiji-recognition \\
 \hline
\end{tabular}
\end{table*}

A Kaggle competition called ``Opening the door to a thousand years of Japanese culture'' was held from July 19 to October 14, 2019.
Whereas the PRMU algorithm competition involved a recognition of single-character images or images containing a few characters, the Kaggle competition tackled the more challenging task of automatically detecting the position of characters on a page of a classical book and correctly recognizing the type of characters.
Of the 44 books in the Kuzushiji dataset described in \ref{ssec:CODH}, 28 books released before the competition were used as training data, and 15 books released after the competition were used as the test data\footnote{One book was eliminated from the competition.}.
The F value, which is the harmonic mean of the precision (the percentage of correct responses among the characters output by the system) and the recall (the percentage of correct responses among the characters in the test data), was used for evaluation.
For approximately 3 months, many international researchers worked on this competition and achieved a practical level of accuracy (F value of greater than 0.9).
There were two typical methods, namely, a single-stage method that applies detection and recognition simultaneously, and a two-stage method that conducts character detection and recognition in stages. Most of the top teams adopted the two-stage method.
The two-stage method applied detectors such as a Faster R-CNN and CenterNet \cite{CenterNet} to detect character regions, and models such as ResNet \cite{ResNet} to recognize individual characters.
As shown in Table \ref{Table:Kaggle}, the method descriptions and implementations of the top teams were published.
We now describe the methods of the top winning teams.

The Chinese team took first place using a straightforward method with a Cascade R-CNN.
Cascade R-CNN can improve the accuracy of object detection by connecting a Faster R-CNN in multiple stages.
High-Resolution Net (HRNet) \cite{HRNet} was used as the backbone network of the Cascade R-CNN.
HRNet utilizes multi-resolution feature maps and can retain high-resolution feature representations without a loss.
The team was able to achieve a high accuracy while maintaining greater simplicity than the methods used by the other teams because the latest techniques were applied, including a Cascade R-CNN and HRNet, which showed the highest levels of accuracy.

The second place team, from Russia, used a two-stage method of detection and classification.
A Faster R-CNN with a ResNet152 backbone was used for detection.
ResNet and ResNeXt \cite{ResNeXt} were used to estimate the type of characters.
Various efforts have been made to improve the accuracy of recognition.
For example, because books of test data are different from books of training data, pseudo labels have also been used to adapt to the environment of an unknown book (author).  
Moreover, a new character class, called a detection error character class, was added to eliminate the detection error at the classification stage.
Finally, the gradient boosting methods LightGBM \cite{LightGBM} and XGBoost \cite{XGBoost} were also used to further improve the accuracy.

The third place team, from Japan, adopted a two-stage method of character detection using a Faster R-CNN and character-type classification using EfficientNet \cite{EfficientNet}.
The team employed several types of data augmentation to increase the number of training data, as shown below.
First, because color and grayscale images were mixed in the training data, they used a random grayscale, which randomly converts images into monochrome during training.
In addition, the training data were augmented using techniques such as combining multiple images by applying mixup \cite{mixup} and random image cropping and patching (RICAP) \cite{RICAP}, and adding some noise to the image by random erasing. 
Because Furigana\footnote{Furigana is made up of phonetic symbols occasionally written next to difficult or rare Kanji to show their pronunciation} is not a recognition target, post-processing such as the creation of a false positive predictor is used for its removal.

The fourth place team, from China, adopted a different method than the other groups; they used a hybrid task cascade (HTC) \cite{HTC} for character detection followed by a connectionist text proposal network (CTPN) \cite{CTPN} for line-by-line character recognition.
One-line images were resized to 32$\times$800, and then fed to a model that recognizes a single line of text.
A convolutional recurrent neural network (CRNN) model was used for line recognition and had a structure with 200 outputs.
A six-gram language model was trained using the KenLM toolkit \cite{KenLM}, and a beam search was applied to decode the CTC output of the model.
The positional accuracy of the CTC output was improved using multitask learning \cite{CTC}, which added an attention loss.
For the data augmentation methods, contrast limited adaptive histogram equalization, random brightness, random contrast, random scale, and random distortion were used. 
A dropout and cutout were applied as the regularization methods.
Although many other teams reported that language models are ineffective, this team reported a slight increase in accuracy.

The fifth place team, from Germany, reported that the one-stage method using CenterNet is consistently more accurate, unlike the two-stage method of the other top teams.
Although, it is common for object detection tasks to deal with approximately 80 classes of objects, such as the MS COCO dataset, this team showed that detection can be achieved without problems even if the number of classes is large. 
They made several modifications to CenterNet, such as creating it from scratch, avoiding the use of an HourglassNet, and using the ResNet50 and ResNet101 structures with a feature pyramid network. 
They also reported that the use of high-resolution images such as 1536$\times$1536 did not show any improvement.

The seventh place team, from Japan, adopted a two-stage method of character region detection using CenterNet followed by ResNet-based character recognition.
For the data augmentation, flipping, cropping, brightness, and contrast were used to create the detector, and cropping, brightness, contrast, and a pseudo-label were used to create a classifier.
The authors attempted to build their own model instead of using a predefined approach.
In general, although predefined models have been designed to provide local features at a high resolution and a wide field of view at a low resolution, it was not necessary to provide a wide field of view at a significantly low resolution for the Kuzushiji detection task.
The team reported that the success of this model was due to the fact that they built their own task-specific model with a high degree of freedom.

The eighth place team, also from Japan, adopted a two-stage method of character detection using CenterNet (ResNet18, UNet) and character classification using MobileNetV3 \cite{MobileNetV3}.
The character detection was made by maintaining the aspect ratio to prevent a separation of characters, and bounding box voting \cite{BBV} was used to reduce the number of undetected characters.
Because the appearance of the characters varies significantly depending on the book applied, the team augmented the data, such as through a grid distortion, elastic transform, and random erasing, to increase the visual variation.

The nineth place team adopted a simple two-stage method using CenterNet with HourglassNet as the backbone for detection and ResNet18 for classification.
Because the detection accuracy was insufficient, multiple results were combined.
For character classification, the team treated characters with less than 5 data as pseudo-labels, $NaN$, because 276 of the characters were not in the training data.

The 13th place team used a Faster R-CNN to detect and classify the character regions, but did not share the network for the detection and classification.
A wide ResNet-34 was used as the backbone of the network for both character detection and character recognition.
The team tried to apply deeper networks, such as ResNet-50 and ResNet-101, but reported that the wider and shallower networks achieve a better accuracy.
They also reported that the use of color images slightly improves the accuracy of detection but worsens the accuracy of the character recognition.
Although they attempted to use a language model to correct incorrect labels, it was ineffective.

The 15th place team, from Japan, used a two-stage method of detection and classification.
First, they applied a grayscale conversion, Gaussian filter, gamma correction, and Ben's pre-processing of the images.
A two-stage CenterNet with HourglassNet as a backbone was applied for character detection.
In the first stage of CenterNet, the bounding box was estimated, and the outside of the outermost bounding box was removed. 
In the second stage of CenterNet, the bounding box was estimated again, and the results of the first and second stages were combined.
Random erasing, horizontal movement, and a brightness adjustment were used to augment the data when creating the detection model.
For character classification, the results of three types of ResNet-based models also trained using pseudo labels were combined and output.
Horizontal movement, rotation, zoom, and random erasing were used for data augmentation when creating the character classification model.

Summarizing the methods of the top teams explained here, we can see that it is important to make use of recently proposed detection and classification models.
However, it is impossible to determine whether a one-stage or two-stage method achieves better results.
For the detection method, models such as YOLO \cite{YOLO}, which are frequently used in object detection, did not perform well, whereas Faster R-CNN and CenterNet were successful.
As the reason for this, there is almost no overlap of characters in the image, which differs from conventional object detection.
Some teams reported an improved accuracy using the latest models with strong backbones, whereas others reported that their own methods were more successful.

\section{Activities Related to Kuzushiji recognition}

The ``Japanese Culture and AI Symposium 2019''\footnote{http://codh.rois.ac.jp/symposium/japanese-culture-ai-2019/} was held in November 2019.
This symposium introduced leading research being conducted on Kuzushiji globally, and the participants discussed the past and present studies with an aim toward future research using AI for the reading and understanding of Kuzushiji.

In the ``Cloud Honkoku (Cloud Transcription)'' project, a system that allows correcting the transcriptions of other participants was implemented.
By cooperating with the Kuzushiji learning application using KuLA and AI technology, the methods developed by the participants were simplified and made more efficient.
The transcribed results can also be used as training data.

CODH provides several online services such as the KuroNet Kuzushiji character recognition service, KogumaNet Kuzushiji character recognition service, and International Image Interoperability Framework (IIIF)\footnote{A set of technology standards intended to make it easier for researchers, students, and the public at large to view, manipulate, compare, and annotate digital images on the web. \url{https://iiif.io/}} compatible Kuzushiji character recognition viewer.
The KuroNet Kuzushiji recognition service\footnote{\url{http://codh.rois.ac.jp/kuronet/}} provides a multi-character Kuzushiji OCR function for IIIF-compliant images.
With this service, we do not need to upload images to the server, but can test our own images.
The IIIF-compatible character recognition viewer provides a single-character OCR function for IIIF-compliant images.
As an advantage of using this viewer, it can be applied immediately while viewing an image, allowing not only the top candidate but also other candidates to be viewed.

In the future, it is expected that Kuzushiji transcription will be further enabled by individuals in various fields, such as those involved in machine learning, the humanities, and the actual application of this historical script.

\section{Future Studies on Kuzushiji Recognition}

In Japan, historical materials have been preserved for more than a thousand years; pre-modern books and historical documents and records have been estimated to number approximately three million and one billion, respectively. 
However, most historical materials have not been transcribed.
Currently, because the Kuzushiji database provided by CODH is mainly limited to documents after the Edo period, it is difficult to automatically transcribe books written prior to this period.
To address this issue, it will be necessary to combine multiple books of various periods owned by multiple institutions and rebuild a large-scale Kuzushiji database.
Looking back further into history, the rules for Kuzushiji increase, and thus it is expected that there will be many character types and character forms that cannot be found in the database.
Therefore, we need a framework for handling unknown character types and character forms that do not exist in the training data.
In addition, as one of the problems in character classification, the number of data for each character class is imbalanced, and it is impossible to accurately recognize characters in a class with an extremely small number of data.

In addition, the quality of the images differs significantly from one book to the next, and the writing styles are completely different depending on the author, which also makes character recognition difficult to achieve.
To this end, it is important to increase the number and variation of training data, to incorporate techniques such as domain adaptation, and improve the database itself.

There is room for further studies on not only individual character-by-character recognition, but also word/phrase/sentence-level recognition based on the surrounding characters and context.
In the Kaggle competition, although many teams added contextual information, such as the use of language models, the accuracy did not significantly improve because there are problems specific to Kuzushiji that differ from those of modern languages.
Therefore, we believe it will be necessary to not only improve machine learning techniques such as deep learning, but also learn rules based on specialized knowledge regarding Kuzushiji.

Based on the current state of transcription, it is important to design an interface that not only allows a complete and automatic recognition of characters, but also allows the user to work effortlessly.
It is desirable to have various functions, such as a visualization of characters with uncertain estimation results, and output alternative candidate characters.

\section{Conclusion}

In this paper, we introduced recent techniques and problems in Kuzushiji recognition using deep learning.
The introduction of deep learning has dramatically improved the detection and recognition rate of Kuzushiji. 
In particular, the inclusion of Kuzushiji recognition in the PRMU algorithm contest and Kaggle competition has attracted the attention of numerous researchers, who have contributed to significant improvements in accuracy.
However, there are still many old manuscripts that need to be transcribed, and there are still many issues to be addressed.
To solve these problems, in addition to improving algorithms such as machine learning, it is necessary to further promote the development of a Kuzushiji database and 
cooperation among individuals in different fields.

\end{document}